\documentclass[10pt,twocolumn,letterpaper]{article}

\usepackage[pagenumbers]{cvpr} % To force page numbers, e.g. for an arXiv version

\usepackage[accsupp]{axessibility}  % Improves PDF readability for those with disabilities.

\usepackage{graphicx}
\usepackage{amsmath}
\usepackage{amssymb}
\usepackage{booktabs}

\usepackage{algorithm}
\usepackage{algorithmicx}
\usepackage{algpseudocode}
\usepackage{siunitx}
\usepackage{enumitem}
\setlist[itemize]{noitemsep}
\usepackage{array}

\usepackage{tabularx}
\usepackage{ragged2e}
\newcolumntype{L}{>{\RaggedRight\arraybackslash}X} % modified 'X' column type

\usepackage[pagebackref,breaklinks,colorlinks]{hyperref}

\usepackage[capitalize]{cleveref}
\crefname{section}{Sec.}{Secs.}
\Crefname{section}{Section}{Sections}
\Crefname{table}{Table}{Tables}
\crefname{table}{Tab.}{Tabs.}

\newcommand{\celegans}{\emph{C.\ elegans}\xspace}

\newcommand{\simm}{\mathord{\sim}}

\def\cvprPaperID{8010} % *** Enter the CVPR Paper ID here
\def\confName{CVPR}
\def\confYear{2023}

\begin{document}

%%%%%%%%% TITLE - PLEASE UPDATE
\title{3D shape reconstruction of semi-transparent worms}

%\author{Thomas~P.~Ilett\thanks{%
%    {\tt\small \{T.Ilett, O.Yuval, T.Ranner, N.Cohen, D.C.Hogg\}@leeds.ac.uk}\\
%    {\textbf{Funding} This work was supported by University of Leeds and EPSRC.
%    % LIBRARY SAYS THIS STATEMENT IS NOT REQUIRED!
%    % For the purpose of open access, the author has applied a Creative Commons Attribution (CC BY) licence to any Author Accepted Manuscript version arising from this submission.
%    }\\
%    {\textbf{Author contributions}
%    % https://credit.niso.org/
%    Conceptualisation, Methodology, Formal analysis, Investigation, Software, Visualisation: TPI.
%    Data curation, Validation: TPI, OY.
%    Writing: TPI (original), all (review and editing).
%    Funding acquisition, Supervision: NC, DCH, TR.
%    $\dagger$ Equal contribution.
%}\\
%    {\textbf{Acknowledgements} 
%     Additional thanks to Matan Braunstein (for help with \cref{fig:overview}), Robert I. Holbrook (data), Felix Salfelder (discussions and data), Lukas Deutz (discussions) and Jen Kruger (proof reading).}\\
%     {\textbf{Data availability} Supplementary movies are available here: \url{https://doi.org/10.6084/m9.figshare.22310650}.}}~%
%  ~~~~~Omer~Yuval\footnotemark[1]~%
%  ~~~~~Thomas~Ranner\footnotemark[1]~%
%  ~~~~~Netta~Cohen\footnotemark[1] ${}^{\dagger}$~%
%  ~~~~~David~C.~Hogg\footnotemark[1] ${}^{\dagger}$\\
%    University of Leeds, Leeds, United Kingdom}

\author{Thomas~P.~Ilett\thanks{%
    {\tt\small \{T.Ilett, O.Yuval, T.Ranner, N.Cohen, D.C.Hogg\}@leeds.ac.uk}\newline
    {\textbf{Funding} This work was supported by University of Leeds and EPSRC.
    % LIBRARY SAYS THIS STATEMENT IS NOT REQUIRED!
    % For the purpose of open access, the author has applied a Creative Commons Attribution (CC BY) licence to any Author Accepted Manuscript version arising from this submission.
    }\newline
    {\textbf{Author contributions}
    % https://credit.niso.org/
    Conceptualisation, Methodology, Formal analysis, Investigation, Software, Visualisation: TPI.
    Data curation, Validation: TPI, OY.
    Writing: TPI (original), all (review and editing).
    Funding acquisition, Supervision: NC, DCH, TR.
    $\dagger$ Equal contribution.
}\newline
    {\textbf{Acknowledgements} 
     Additional thanks to Matan Braunstein (for help with \cref{fig:overview}), Robert I. Holbrook (data), Felix Salfelder (discussions and data), Lukas Deutz (discussions) and Jen Kruger (proof reading).}\newline
     {\textbf{Data availability} Supplementary movies are available here: \url{https://doi.org/10.6084/m9.figshare.22310650}.}}~%
  ~~~~~Omer~Yuval\footnotemark[1]~%
  ~~~~~Thomas~Ranner\footnotemark[1]~%
  ~~~~~Netta~Cohen\footnotemark[1] ${}^{\dagger}$~%
  ~~~~~David~C.~Hogg\footnotemark[1] ${}^{\dagger}$\\
    University of Leeds, Leeds, United Kingdom}

\makeatletter
\let\@oldmaketitle\@maketitle% Store \@maketitle
\renewcommand{\@maketitle}{\@oldmaketitle% Update \@maketitle to insert...
	\captionsetup{type=figure}
    \centering
    \includegraphics[width=0.99\linewidth]{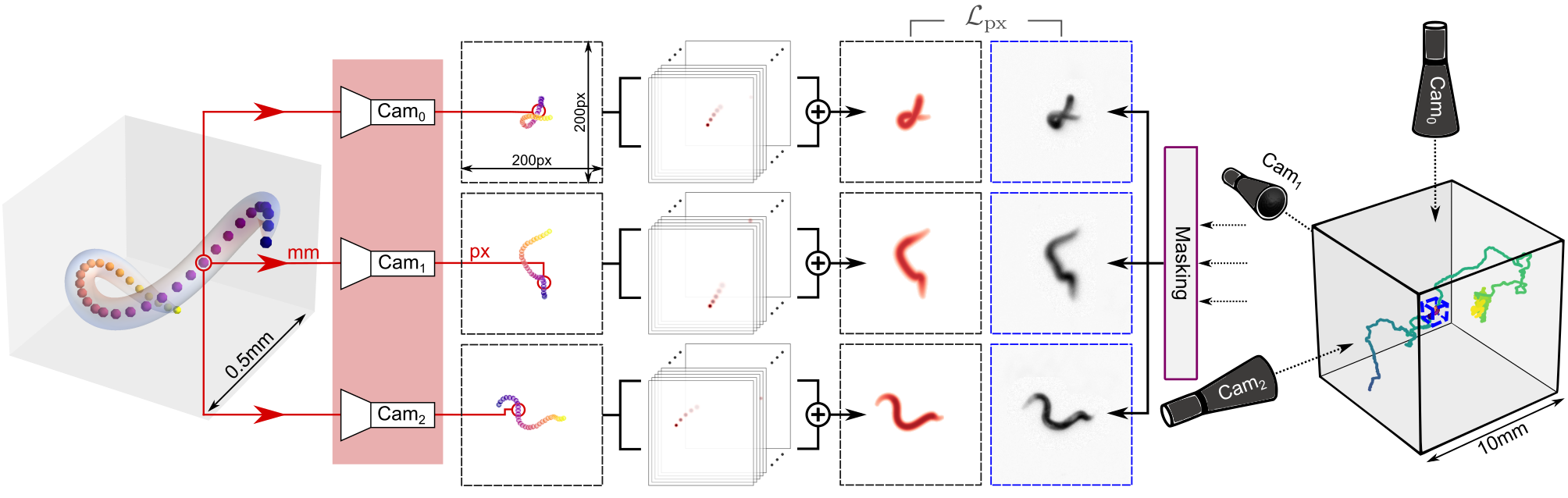}
	\captionof{figure}{Posture reconstruction pipeline and imaging setup.}
	\label{fig:overview}
	\bigskip}
\makeatother

\maketitle

%%%%%%%%% ABSTRACT

\begin{abstract}
3D shape reconstruction typically requires identifying object features or textures in multiple images of a subject. This approach is not viable when the subject is semi-transparent and moving in and out of focus. Here we overcome these challenges by rendering a candidate shape with adaptive blurring and transparency for comparison with the images. We use the microscopic nematode \emph{Caenorhabditis elegans} as a case study as it freely explores a 3D complex fluid with constantly changing optical properties. We model the slender worm as a 3D curve using an intrinsic parametrisation that naturally admits biologically-informed constraints and regularisation. To account for the changing optics we develop a novel differentiable renderer to construct images from 2D projections and compare against raw images to generate a pixel-wise error to jointly update the curve, camera and renderer parameters using gradient descent. The method is robust to interference such as bubbles and dirt trapped in the fluid, stays consistent through complex sequences of postures, recovers reliable estimates from blurry images and provides a significant improvement on previous attempts to track \celegans in 3D. Our results demonstrate the potential of direct approaches to shape estimation in complex physical environments in the absence of ground-truth data.
\end{abstract}

%%%%%%%%% BODY TEXT
\section{Introduction}
\label{sec:intro}

Many creatures such as fish, birds and insects move in all directions to search and navigate volumetric environments. Acquiring 3D data of their motion has informed models of locomotion, behaviour and neural and mechanical control \cite{Holbrook2011,Belant2019}. While technological advances have made the collection of large quantities of multi-viewpoint visual data more attainable, methods for extracting and modelling 3D information remain largely domain-dependant as few species share common geometric models or exist within the same spatial and temporal scales \cite{Partridge1980,Zhu2007,Theriault2010,Simpfendorfer2012,Cooper2014,Macri2017,Ferrarini2018,Berlinger2021,Jiang2022}. Furthermore, while humans and some domesticated animals \cite{Wilhelm2015,Kearney2020} may act naturally while wearing special markers, marker-less observations of many species makes feature extraction more challenging and means pose estimation generally lacks ground-truth data \cite{Sellers2014}.

As a case study in marker-less 3D shape reconstruction, we consider \celegans, a hair-thick, $\simm \SI{1}{\milli\metre}$ long animal with a simple tapered cylinder shape, which can be constructed from a midline ``skeleton''. In the wild, \celegans can be found in a wide range of complex 3D environments, \eg decomposing organic matter, with continually changing physical properties \cite{Felix2010,Frezal2015,Schulenburg2017}. However, to date, experiments have focused nearly exclusively on locomotion on a plane, limiting insight to the constrained, planar behaviours.

We obtained a large dataset (4 hours 53 minutes $\simeq$ 440,000 frames at 25Hz) of experimental recordings of individual worms moving freely inside a glass cube filled with a gelatin solution. The cube is positioned between three nearly-orthogonal static cameras fitted with telecentric lenses. Initial pinhole camera model parameter estimates are provided \cite{Salfelder2021} but are imprecise and require continuous adjustment across the course of a recording to account for small vibrations and optical changes to the gel. We aim to simultaneously reconstruct a 3D shape and find corrected camera parameters to match these recordings in a process akin to bundle adjustment \cite{Triggs1999}.

3D reconstruction typically involves the identification and triangulation of common features from multiple viewpoints or the synthesis of full images including texture and shading information to match given scenes \cite{Hartley2003,Seitz2006,Zollhofer2018,Fu2021}. Imaging animals with length $\simm \SI{1}{\milli\metre}$ requires sufficient magnification, but simultaneously capturing long-term trajectories up to 25 minutes requires a large volume of view (10-20 worm lengths per axis). As the worm explores the cube it frequently appears out of focus in one or more of the cameras. Air bubbles and dirt trapped in the gel along with old tracks are difficult to differentiate from the transparent worm, particularly at the tapered ends. Self occlusion invariably appears in a least one view, where hidden parts darken the foreground while the ordering of fore/back-parts is not discernible. As the semi-transparent and self-occluding subject moves in the volume, photometric information in one view bears little relevance to the appearance in the others making feature identification and photometric matching particularly challenging. We found that standard approaches may suffice for limited sub-clips, but lose parts of the object or fail catastrophically for much of the data and the solution requires a degree of adaptation.

We present an integrated ``project-render-score'' algorithm to obtain a midline curve for each image-triplet (\cref{fig:overview}). Discrete curve vertices are \emph{projected} through a triplet of pinhole camera models, \emph{rendered} to produce an image-triplet for direct comparison against the recorded images and \emph{scored} according to their intersection with worm-like pixels in all three views. The differentiable renderer stacks 2D super-Gaussian blobs at the projected locations of each vertex to approximate the transparency along the worm, accounting for the variable focus and providing soft edges that direct the geometric model towards the midline. The scoring allows the detection of incongruities and keeps the curve aligned to the worm in all views. Regularisation terms ensure smoothness along the body and in time. Curve, camera and rendering parameters are jointly optimised using gradient descent to convergence. Once the worm shape has been resolved, it is generally only lost during image degradation or significant self-occlusions that make the posture unresolvable by eye. 

In summary, our main contributions are:
\begin{itemize}[topsep=0pt]
	\item A robust pipeline for 3D posture reconstruction of a freely deforming semi-transparent object from noisy images.
	\item A novel viewpoint renderer to capture optical distortions and transparency.
	\item A feature-free bundle adjustment algorithm using direct image comparison and gradient descent.
\end{itemize}

%

%------------------------------------------------------------------------
\section{Related work}

\paragraph{Bundle adjustment (BA)}
\hspace{-1em}
is a procedure to jointly optimise 3D geometry and camera parameters \cite{Triggs1999,Hartley2003}. BA typically identifies common features of an object from multiple viewpoints in order to minimise a prediction error between projections of the corresponding 3D points and their 2D observations. BA is frequently used in conjunction with other methods to find camera parameters using multiple images of a 3D calibration object with known control points or for fine-tuning results \cite{Longuet-Higgins1981,Tsai1987,Weng1992,Huang1995,Faugeras2001,Ozyecsil2017}. 

Feature detection converts photometric information into image coordinates. In BA, coordinates of common features are used to solve a geometric optimisation problem. Photometric bundle adjustment methods additionally require objects to have the same appearance in all views \cite{Georgel2008,Delaunoy2014}. Our method is entirely photometric, as such differing from BA. As our objects appear differently across views, all pixel information is used and the geometry is solved intrinsically.

\paragraph{Pose estimation}

Deep network approaches have proved well-suited to 2D human-pose estimation as they are potent feature extractors and large annotated training sets are available \cite{Andriluka2014,Toshev2014,Sun2019}. For 3D postures, ground truth multi-view datasets are less common. Recent progress \cite{Liu2022} relies on end-to-end architectures \cite{Tekin2016,Pavlakos2017,Guler2018,Kanazawa2018,Kolotouros2019,Wu2020} or splitting the problem into 2D pose estimation and then constructing the 3D pose \cite{Martinez2017,Chen2017a}. 
Despite similar approaches used for non-human pose estimation, the huge variability in scales and shapes among species introduces a variety of challenges \cite{Jiang2022}. Motion capture in controlled settings with markers (providing ground truth skeleton and joint angle data for humans, horses and dogs \cite{Wilhelm2015,Kearney2020}), are not available for most animals. Generalised mesh surfaces may be used, but often require multiple views and thousands of parameters, and do not guarantee consistency through time. In contrast, approximating an animal shape using a few-parameter morphable model can be both tractable and robust. Successful examples include swimmers \cite{Prasad2010,Cashman2012}, birds \cite{Vicente2013,Kanazawa2018}, mammals \cite{Bala2020,Biggs2020,Kanazawa2016,Ntouskos2015} and generic quadrupeds \cite{Zuffi2017,Biggs2019}. However, these methods expect opaque subjects with consistent textural appearances between views. 

\celegans has a simple geometric shape that can be well reconstructed from a midline skeleton and parametrised by curvature values along the body (see \cref{sec:geometric_model}). This is the deformable template we look to fit to the data. Despite the apparent simplicity, each vertex of the discretised curve has two degrees of freedom (two curvature values) and as we use 128 vertices, our model is highly deformable and requires many parameters (although smoothness regularisation simplifies the problem somewhat). In contrast to deep-learning approaches, our model includes only a small number of explainable parameters and direct optimisation avoids lengthy training and dataset requirements.

\paragraph{\celegans}

Numerous freely available software packages are capable of simultaneous tracking and skeletonising single or multiple worms in 2D using inexpensive microscopic imaging \cite{Swierczek2011,Yemini2011,Ramot2008,Javer2018,Sznitman2010,Berri2009} (see \cite{Husson2018} for a review). Most of these skeletonisers combine image segmentation to separate the animal from the background with thinning of the mask to some midline pixels and fitting a spline.

The 3D reconstruction problem has received relatively little attention. Using at first two views \cite{Kwon2013} and then three, Kwon \etal \cite{Kwon2015} designed a motorised stage coupled with a real-time tracker to keep a worm in focus under high magnification in a 3D environment while capturing trajectories of up to 3 minutes. Thresholded images are lifted into 3D, intersected in voxel space and thinned \cite{Guo1989} to produce a final skeleton. Kwon \etal omit camera modelling and assume perfectly parallel projections -- assumptions that result in large errors for the data we use. Shaw \etal \cite{Shaw2018} employed light field microscopy to generate depth maps alongside images from a single viewpoint. A midline skeleton is generated by fitting a spline to the 3D coordinates of the central voxels. However, self-occlusions cannot be resolved and only relatively planar postures were investigated. 

Salfelder \etal \cite{Salfelder2021} and Yuval \cite{Omer2022} both present 3D reconstruction algorithms using the three-camera set up and calibration described in \cite{Salfelder2021}. In Salfelder \etal \cite{Salfelder2021}, a neural network is trained to identify 2D midlines from individual camera images before lifting into 3D voxel space. To account for changing camera parameters, a relative axial shift $(dx,dy,dz)$ is optimised for each frame-triplet to maximise the voxel intersection before thinning. Remaining voxel coordinates are used as control points to fit a curve using a finite-element formulation. This approach works well when the midline is well detected in each of the views, but can fail on occluded postures or low-resolution, blurry images.

Yuval \cite{Omer2022} uses a neural network to track head and tail points in 3D lab coordinates and a curve is fit between these fixed end points using a hill-climbing optimisation algorithm. Scoring is based on curve smoothness and pixel intensities at the projected curve points. This method works well when the head and tail are correctly identified but struggles, or requires manual correction, otherwise.

In our approach we find that incorporating the camera model parameters into the optimisation results in more robust and accurate results. This extends the idea proposed in Salfelder \etal \cite{Salfelder2021} that adjusting the relative positions of the cameras could result in large gains in accuracy. It is likely that the relative shift adjustments, presented there, account for the changing optical properties.

%------------------------------------------------------------------------

\begin{figure*}
	\centering
	\includegraphics[width=0.9\linewidth]{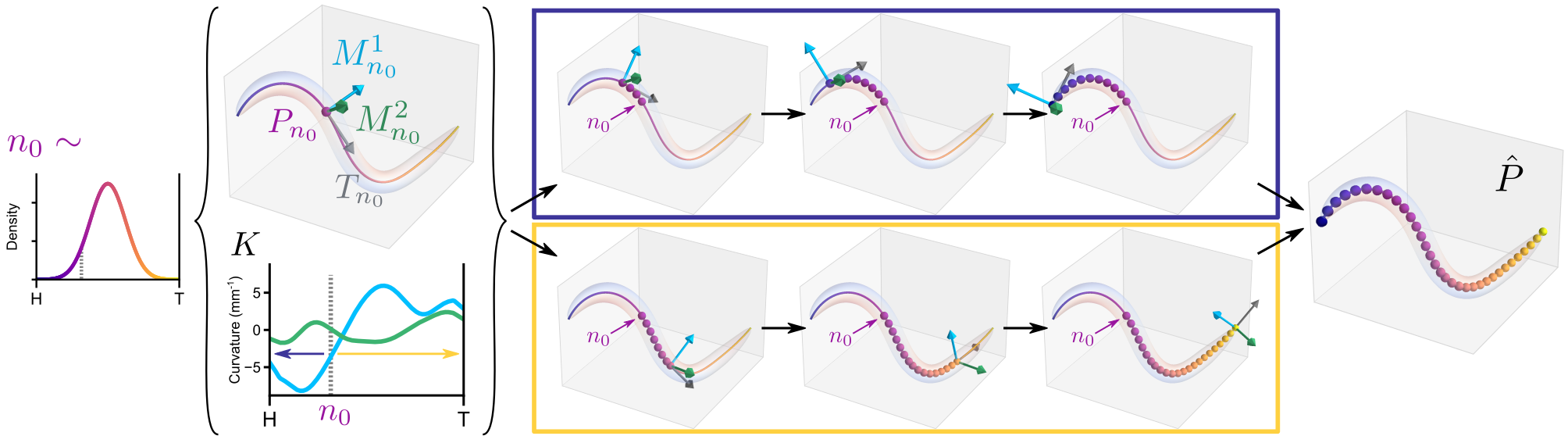}
	\caption{The 3D curve is traced out from initial point $P_{n_0}$ and orientation frame $(T_{n_0},M^1_{n_0},M^2_{n_0})$. The index $n_0$ of the initial point is drawn from a normal distribution at each iteration to prevent kinks developing through repeated use of the same starting point. The final curve $\hat{P}$ is computed in two parts by integrating the Bishop equations with curvature $K$ towards the head and tail separately.}
	\label{fig:curve}
\end{figure*}

\section{Geometric model}\label{sec:geometric_model}

Nematode shapes can be well approximated by a tapered cylinder and computed from a midline. We construct the midline curve in 3D using an object-centric parametrisation, separating shape from position and orientation to allow us to easily constrain and regularise the shape to stay within biologically-reasonable bounds. We discretise the curve into $N$ equidistant vertices and encode the posture in curvature $K \in \mathbb{R}^{N \times 2}$ and length $l \in \mathbb{R}$ that fully define the shape up to a rigid-body transformation.

We express the 3D curve using the Bishop frame \cite{Bishop1975}, given by $TM^1M^2$ where $T$ is the normalised tangent of the curve and $M^1,M^2$ form an orthogonal basis along the midline. At vertex $n$, the curvature is $K_n=(m^1_n, m^2_n)$, where $m_n^1,m_n^2 \in \mathbb{R}$ are the curvature components along $M^1, M^2$. (The more familiar Frenet frame is less stable as it is undefined at zero-curvature points.) Numerical integration of a system of difference equations from starting point $P_{\mathrm{init}}$ and initial orientation $(T_{\mathrm{init}},M^1_{\mathrm{init}},M^2_{\mathrm{init}})$ yields the curve path $P \in \mathbb{R}^{N \times 3}$. See \cref{app:bishop_frame} for details.

During optimisation, errors accumulate near the starting point, $P_{\mathrm{init}}$, resulting in either parts of the curve moving faster than other or kinks developing (even with strong regularisation).
To resolve this we sample an initial vertex index $n_0$ from a Gaussian distribution (subject to rounding) centred at the middle index at every optimisation step.
Setting the starting point $P_{\mathrm{init}}=P_{n_0}$ has the effect of continually shifting the discontinuity so kinks are never given the opportunity to develop (\cref{fig:curve}). Summarising the integration as $F$, the 3D curve is generated from the parameters:
\begin{align}
(\hat{P},\hat{T},\hat{M}^1) &= F\left(P_{n_0},T_{n_0},M^1_{n_0},K,l,n_0\right).  \label{eq:curve}
\end{align}

Each gradient update adjusts all curvature values $K$ but the position and orientation only at the randomly selected $n_0$ vertex $(P_{n_0},T_{n_0},M^1_{n_0})$. Updating $(P,T,M^1)$ at only this vertex produces a $P$ that is inconsistent with the updated $K$. Therefore, after applying gradient updates we re-compute the full curve and orientation from $n_0$ and set $(P,T,M^1)$ to the output $(\hat{P},\hat{T},\hat{M^1})$.

Since the curve describes a biological creature, we constrain the length $l$ to $(l_\text{min},l_\text{max})$ and limit the curvature by $|K_n| < 2\pi k_\text{max}$. The values of $(l_\text{min},l_\text{max})$ we use vary depending on magnification but the bounds do not need to be tight and are in the range $0.5$--$\SI{2}{\milli\metre}$.
The curvature constraint $k_{\text{max}}$ is set by considering the number of circle achieved by a constant curvature curve and is fixed at $3$.

%------------------------------------------------------------------------

\section{Project, Render, Score}\label{sec:prs}

The core of the optimisation pipeline is separable into three main stages; project, render and score. The 3D curve $\hat{P}$ generated in \cref{eq:curve} is \emph{projected} through the camera models into 2D points that are \emph{rendered} into images and then \emph{scored} against the three views.

\subsection{Project} \label{sec:project}

The cameras are modelled using a triplet of pinhole camera models with tangential and radial distortion that project 3D points into image planes using perspective transformations (see \cref{app:pinhole_cams}). Each pinhole camera model offers a simple (15 parameters, $\{\eta_c\}$), tractable, approximation to the optical transformation. We also include relative shifts along the local coordinate axes, $\eta^s=(dx,dy,dz)$, shared between the three models, as proposed by Salfelder \etal \cite{Salfelder2021}. Initial camera coefficients for the triplet-model are provided along with the recordings and typically give root mean squared reprojection errors up to 10 pixels ($\sim \mathcal{O}(\text{worm radius})$). 

Due to the initial calibration errors and changes in optical properties as the gelatin sets and is disturbed by the worms we re-calibrate the cameras at every frame by including the camera parameters in the optimisation step. To avoid an under-determined problem, after we have found a configuration that supports good reconstructions for a recording we fix all but the $\eta^s$ parameters. Interestingly, we still see changes (up to $30\si{px} \sim \SI{0.15}{\milli\metre}$) in $\eta^s$ but as this relates to the relative positioning it does not affect the posture reconstruction or long-term trajectories.

Projecting the 3D curve $\hat{P}$ through the camera-triplet model $\Gamma$ with parameters $\eta = \lbrace \eta_0, \eta_1, \eta_2, \eta^s \rbrace$ generates 2D image points per view, which we combine as $Q = \Gamma(\hat{P}, \eta) \in \mathbb{R}^{3 \times N \times 2}$.

\subsection{Render} \label{section:render}

\begin{figure}
	\centering
	\includegraphics[width=0.9\linewidth]{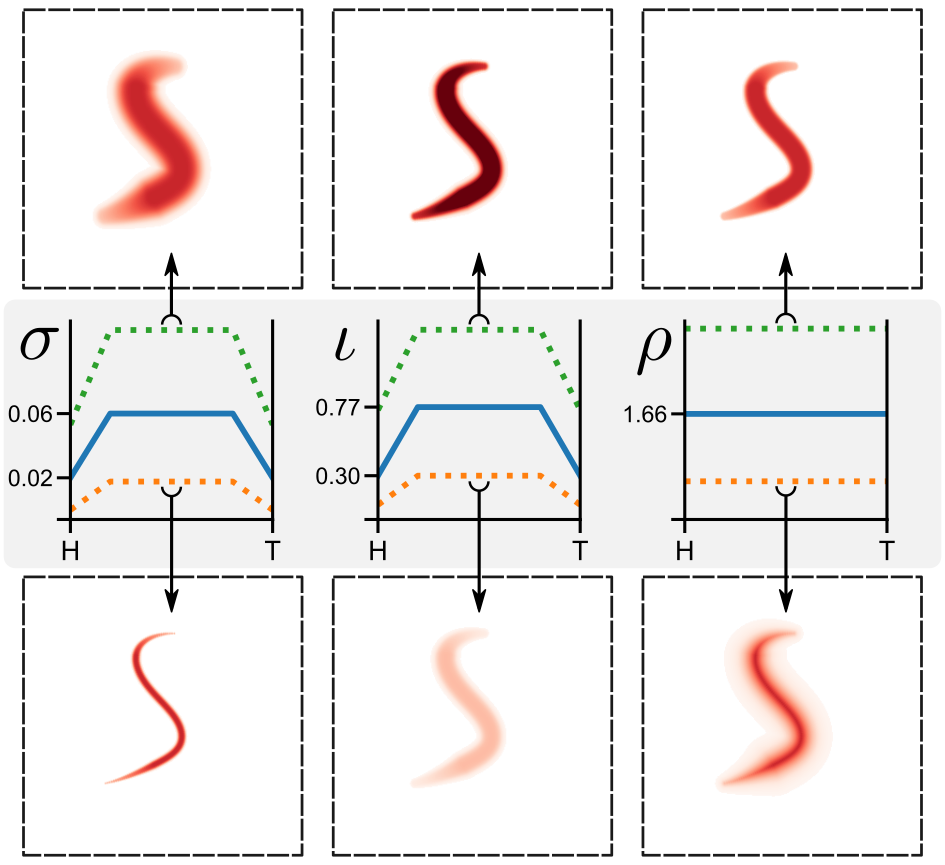}
	\caption{The rendering stage generates super-Gaussian blobs at each vertex position on the image. The shape of the blobs depends on the optimisable parameters: the scale $\sigma$, the intensity $\iota$ and the exponent used in the Gaussian $\rho$. $\sigma$ and $\iota$ are tapered down to fixed minimum values at the head and tail. The effects of varying these parameters from a converged solution (blue curves) are shown above (green curves) and below (orange curves) each.}
	\label{fig:render}
\end{figure}

In order to evaluate the reconstruction directly against the raw data, we render the projected 2D midline points into images using optimisable shape and rendering parameters. Since worm bodies are well approximated by tapered cylinders, in theory we only require maximum and minimum radius values and a tapering function. However, \celegans are semi-transparent -- increasingly so at the head and tail -- and their internal anatomy has varying optical properties that diffract and distort the light. These challenges are further exacerbated by the worms often being out of focus in at least one of the views, therefore even an anatomically accurate model stands little chance of being correctly resolved.

We render realistic images by combining 2D super-Gaussian functions centred on each projected vertex. Crucially, we allow the rendering parameters to differ between cameras since the animal seldom has the same photometric qualities in different views. We optimise three parameters for each camera view $c$: $\sigma_c \in \mathbb{R}$ controls the spread, $\iota_c \in \mathbb{R}$ scales the intensity, and $\rho_c \in \mathbb{R}$ sharpens or softens the edges (\cref{fig:render}). To capture the tapered shape we weight $\sigma_c$ and $\iota_c$ from their optimisable values along the middle $60\%$ to minimum values $\sigma_{\text{min}}$ and $\iota_{\text{min}}$ at the ends and define the tapered outputs $\bar{\sigma}_c \in \mathbb{R}^N$ and $\bar{\iota}_c \in \mathbb{R}^N$ (\cref{app:render_params}). $\sigma_{\text{min}}$ and $\iota_{\text{min}}$ are manually fixed for each recording to account for different magnification factors and worm size variability. 

For each camera index $c$ and vertex index $n$ we define the rendered blob $B_{c,n} \in \mathbb{R}^{w \times w}$ (image size $w$) for pixel $(i,j)$ as:
\begin{equation} \label{eq:blobs}
B_{c,n}(i,j) = \bar{\iota}_{c,n}
				\exp{
                    \left[
    					-\left(
    						\frac{(i - Q_{c,n,0})^2 + (j - Q_{c,n,1})^2}
    								{2\bar{\sigma}_{c,n}^2}
    					\right)
    					^{\rho_c}
                    \right]
				}.
\end{equation}
The stacks of blobs are combined to generate the complete renderings $R \in \mathbb{R}^{3 \times w \times w}$ by taking the maximum pixel value across all blobs: for pixel $(i,j)$,
\begin{equation} \label{eq:blobs_render}
R_c(i,j) = \max \left\lbrace B_{c,n}(i,j) \right\rbrace_{n=0,\dots,N-1}.
\end{equation}

The orientation of the body directly affects the pixel intensity of both raw and rendered images. When pointing directly at a camera the peaks of the blobs cluster closely together and appear as a high-intensity (opaque) circle. Pointing laterally causes the peaks to spread out on the image revealing more of the lower-intensity tails. In both situations our blob-rendering approach approximates transparency effects in the raw images without the need to model complex intensity-orientation responses. Moreover, super-Gaussian blobs allow sharp outlines to be produced in one view by using a large exponent and flat-top blobs, and blurry images to be produced for another, using low intensity and high variance.

\subsection{Score}

\begin{figure}
	\centering
	\includegraphics[width=0.95\linewidth]{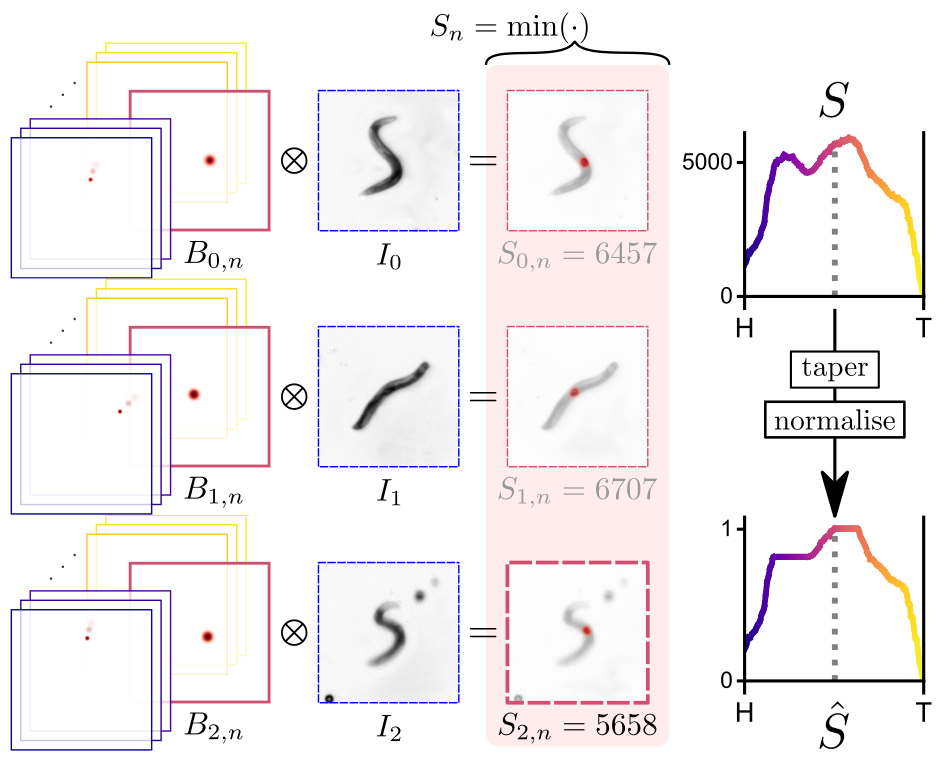}
	\caption{The 3D curve points are scored individually according to how well they match the three views. The triplet of blobs associated with vertex $n$ ($B_{.,n}$) are multiplied with the images $I$ and summed. We take the minimum of the three sums and then taper these values from the midpoint-out.}
	\label{fig:score}
\end{figure}

In order to evaluate how well the curve represents the worm we require a way of distinguishing between worm-pixels and non-worm pixels such as dirt, bubbles, old tracks and even other worms. When the animal truly intersects with environmental interference it can be impossible to differentiate between the two, but in the majority of cases there exists a gap between the worm and the noise that is visible in at least one of the views. By ensuring that the curve corresponds to a single contiguous pixel mass in \emph{all} of the images we are able to safely ignore other artefacts (\cref{fig:score}).

To detect if the curve is bridging a gap, each vertex $\hat{P}_n$ is scored by correlating its corresponding blobs $B_{.,n}$ (\cref{section:render}) with the images $I$. The raw score $S_n \in \mathbb{R}$ is defined:
\begin{equation}
S_n = \min \left\lbrace \frac{\sum_{i,j} B_{c,n} \cdot I_c}{\bar{\sigma}_{c,n} \bar{\iota}_{c,n}} \right\rbrace_{c=0,1,2}
\end{equation}
where $\cdot$ is element-wise multiplication and the sum is taken over the image dimensions. By taking the minimum we ensure that vertices failing to match pixels in any one of the views will receive low scores regardless of how well they match pixels in the other views.

If the curve is bridging two disjoint groups of pixels that are visible in all three views this will present as two peaks in $S$. Since we are only interested in finding one object we restrict the scores to contain just one peak by tapering $S$ from the middle-out to form the intermediate $S'$. Finally we normalise $S'$ to get scores $\hat{S}$ relative to the peak:
\begin{align}
S'_n &= 
\begin{cases}
	\min \lbrace S_n, S'_{n+1} \rbrace		& 0 \leq n < N/2 \\
	S_n										& n = N/2 \\
	\min \lbrace S_n, S'_{n-1} \rbrace		& N/2 < n < N
\end{cases} \\
\hat{S} &= \frac{S'}{\max_n \lbrace S' \rbrace}.
\end{align}

The final score profile $\hat{S}$ provides insight into how well the curve matches a contiguous pixel mass across all three views and how evenly that mass is distributed.

\paragraph{Masking}

\begin{figure}
	\centering
	\includegraphics[width=0.95\linewidth]{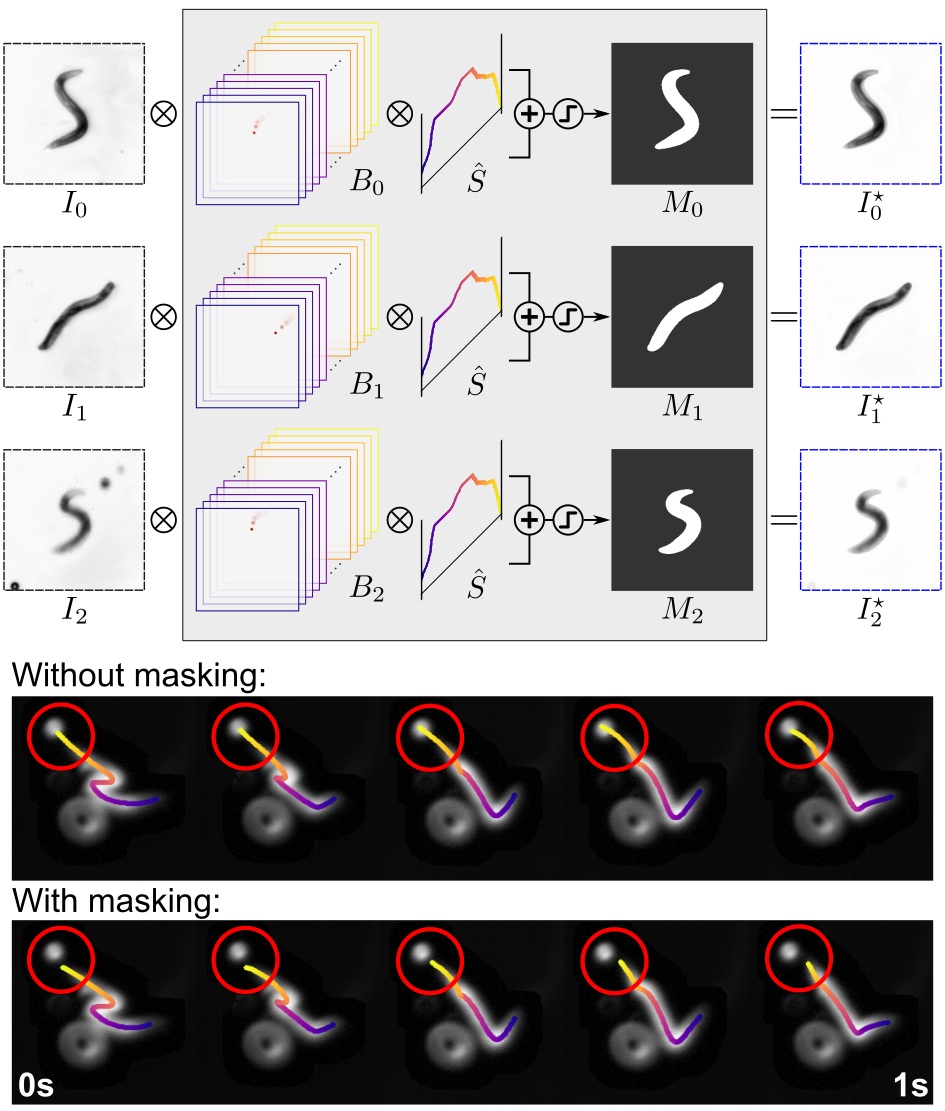}
	\caption{The noisy input images are cleaned by applying masks that force pixel-errors to be local to the current estimate. The blobs $B$ are scaled by the relative scores $\hat{S}$, combined using the maximum pixel value across blobs and thresholded to form the masks $M$. The masks are applied to the raw input images $I$ to generate the targets: $I^\star$. Masking ensures only a single contiguous pixel mass is detected. Without it, parts of the reconstruction can ``stick'' to nearby bubbles and other artefacts as shown below.}
	\label{fig:masks}
\end{figure}

From the score profile $\hat{S}$ we identify image areas that are more likely to contain the pixel masses that correspond to the worm. Masks $M \in \mathbb{R}^{3 \times w \times w}$ applied to the input, $I^\star = M \cdot I$, focuses attention (and gradient) to only these areas of interest, consistently across all three views and exclude interference outside the masks (\cref{fig:masks,app:masks}). Pixel intensities outside the masks are significantly reduced, but not zeroed in order to avoid stagnation in case the reconstruction completely misses the worm.

\paragraph{Centre-shifting} \label{sec:curve-shift}

\begin{figure}
	\centering
	\includegraphics[width=0.95\linewidth]{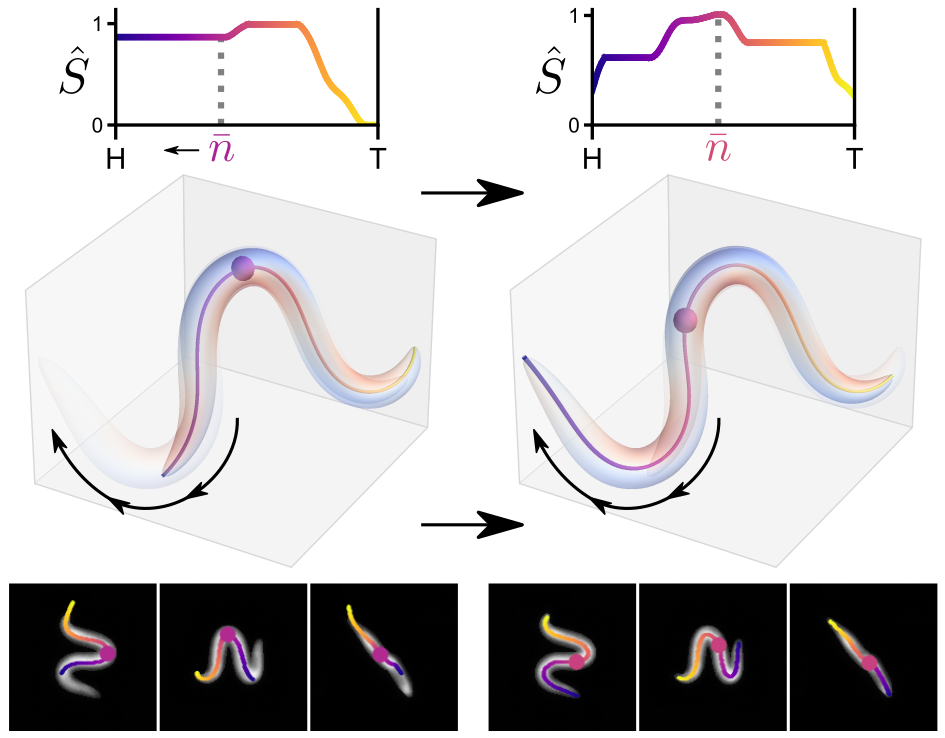}
	\caption{As the animal moves along the path of its midline the tail may be left behind (left column). This can be identified from an unbalanced score profile $\hat{S}$. By periodically shifting the curve along its length (adding new curvature values at one end and discarding from the other) the centroid index ($\bar{n}$) of the scores can be centred. Gradient descent optimisation then updates the new curvature values so the curve matches the target (right column).}
	\label{fig:centre_shift}
\end{figure}

The scores $\hat{S}$ also indicate the relative positioning of the curve over the target object. As the curve aligns with a pixel mass, vertices with high scores (apparently ``converged'') tend to lock into place thus hindering convergence of the rest of the object. For each frame, we use the previous frame solution as the starting point, so the majority of points rapidly converge. However, errors introduced at the tips remain as they are insufficient to generate the collective shift required. The effect can easily be identified from an unbalanced score profile (\cref{fig:centre_shift}) and rectified by periodically shifting the curve along its length between gradient descent optimisation steps (\cref{app:centre_shift}).

%------------------------------------------------------------------------

\section{Optimisation}

The main pixel-loss to be minimised is defined as:
\begin{equation}
\mathcal{L}_\text{px} = \frac{1}{3 w^2} \sum_{c,i,j} (R_c(i,j) - I^\star_c(i,j))^2.
\end{equation}
To improve head and tail detection we also minimise a scores-loss,
\begin{align}
\mathcal{L}_\text{sc} &= \frac{\max(S')N}{\sum_n S''_n}, \text{ where} \\
S''_n &= S'_n \left( \frac{2n-(N-1)}{N-1} \right)^2,
\end{align}
that is quadratically weighted towards the tips where the scores are naturally lower due to the transparency.

In addition we include a number of regularisation terms. To keep the curve smooth we define
\begin{equation} \label{eq:L_smooth}
\mathcal{L}_\text{sm} = \sum_{n=1}^{N-1} | K_n - K_{n-1} | ^2,
\end{equation}
where $|\cdot|$ is the $l^2\text{-norm}$. To ensure all parameters change smoothly between frames we set
\begin{equation}
\mathcal{L}_\text{t} = \sum_{x \in \lbrace l, K, \hat{P}, \eta, \sigma, \iota, \rho \rbrace} |x^\text{prev} - x|^2,
\end{equation}
where $x^\text{prev}$ refers to the frozen value of the variable from the previous frame. And to avoid self-intersections, we use
\begin{align}
d_{n,m} &= |\hat{P}_n - \hat{P}_m|, \\
d'_{n,m} &= \frac{1}{3} \sum_c \bar{\sigma}_{c,n} + \frac{1}{3} \sum_c \bar{\sigma}_{c,m}, \text{ and } \\
% n_d &= N/k_\text{max}, \\
\mathcal{L}_\text{i} &= \sum_{n=0}^{N-N/k_\text{max}-1} \sum_{m=n+N/k_\text{max}}^{N-1} \begin{cases} \frac{d'_{n,m}}{d_{n,m}},  & \text{if } d_{n,m} < d'_{n,m} \\ 0, & \text{otherwise.} \end{cases}   \label{eq:Li}
\end{align}
A loss is incurred, $\mathcal{L}_{\text{i}} > 0$, when two points which are sufficiently far apart ($> N/k_\text{max}$) along the curve come within a distance defined by the sum of their mean rendering variances (since these approximate the worm's radius). \cref{eq:Li} forces the algorithm to find postures that are always feasible even during self-occlusions and complex manoeuvres.

The losses are combined in a weighted sum to yield the final optimisation target:
\begin{align} \label{eq:loss}
\mathcal{L} = \omega_\text{px}\mathcal{L}_\text{px} 
				+ \omega_\text{sc}\mathcal{L}_\text{sc} 
				+ \omega_\text{sm}\mathcal{L}_\text{sm}
				+ \omega_\text{t}\mathcal{L}_\text{t}
				+ \omega_\text{i}\mathcal{L}_\text{i}.
\end{align}
Values of $\omega$ used in our experiments are listed in \cref{tab:omegas}.

To achieve robust reconstructions it is important that the curve parameters learn fastest, then the rendering parameters and finally the camera parameters. Imposing this hierarchy of rates ensures camera model stability and prevents the renderer from over-blurring the edges (as it tries to ``reach'' the pixels). Thus, movement between frames is primarily captured through curve deformations. We use learning rates $\lambda_p=\num{1e-3}$ for the curve parameters $\lbrace P, T, M^1, K, l \rbrace$, $\lambda_r=\num{1e-4}$ for the rendering parameters $\lbrace \sigma, \iota, \rho \rbrace$ and $\lambda_\eta=\num{1e-5}$ for the camera parameters $\eta$.

The curve is initialised as a small ($\simm \SI{0.2}{\milli\metre}$), randomly oriented straight line centred in the field of view of all three cameras. We slowly increase the length to $l_\text{min}$ over the first $200\text{-}500$ steps as the curve gets positioned and orientated.

The pipeline is constructed using PyTorch \cite{paszke2017automatic} and the loss minimised is using Adam \cite{Kingma2014} with periodic centre-shifting of the curve vertices. Learning rates are decreased by a factor of $0.8$ for every $5$ steps taken without improvement in $\mathcal{L}$ to a minimum of $\num{1e-6}$ until convergence is detected. Subsequent frames are instantiated with the solution from the previous frame for efficiency and to maintain consistency through complex sequences of self-occluding postures. Example videos showing the effects of varying some of the options on the optimisation are described in \cref{app:results}.

%------------------------------------------------------------------------

\section{Results}

\begin{figure}
	\centering
	\includegraphics[width=0.9\linewidth]{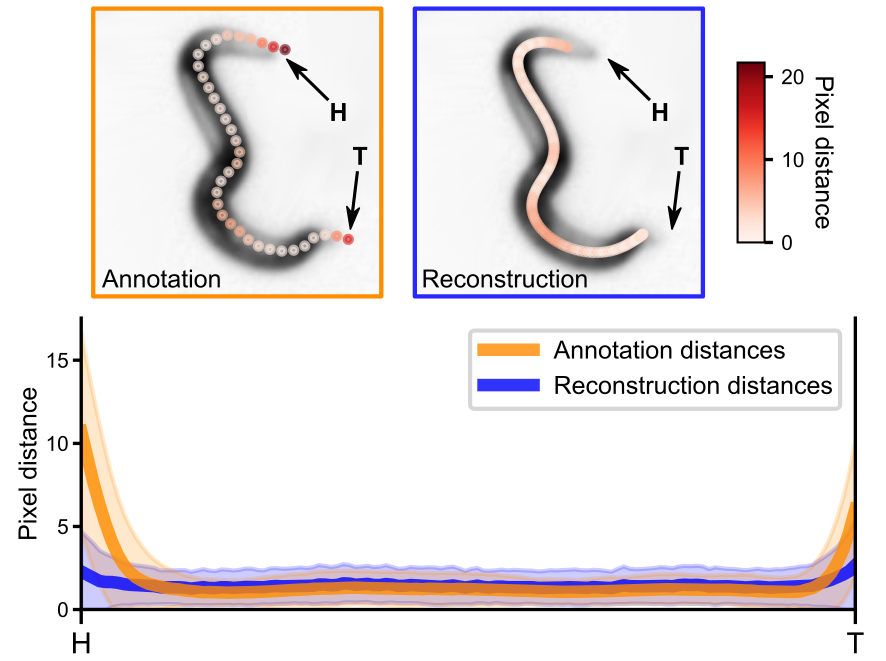}
	\caption{Validation against 487 manual annotations. At the top we show an example of an annotated frame (left, orange) alongside a projection of our matching 3D midline (right, blue). Below we plot the sample averages $\pm 2 \text{std}$. We find our midlines are consistently close to annotated points (blue curve), but annotations typically extend further into the head and tail regions (orange curve).}
	\label{fig:validation}
\end{figure}
\begin{figure*}
	\centering
	\includegraphics[width=0.95\linewidth]{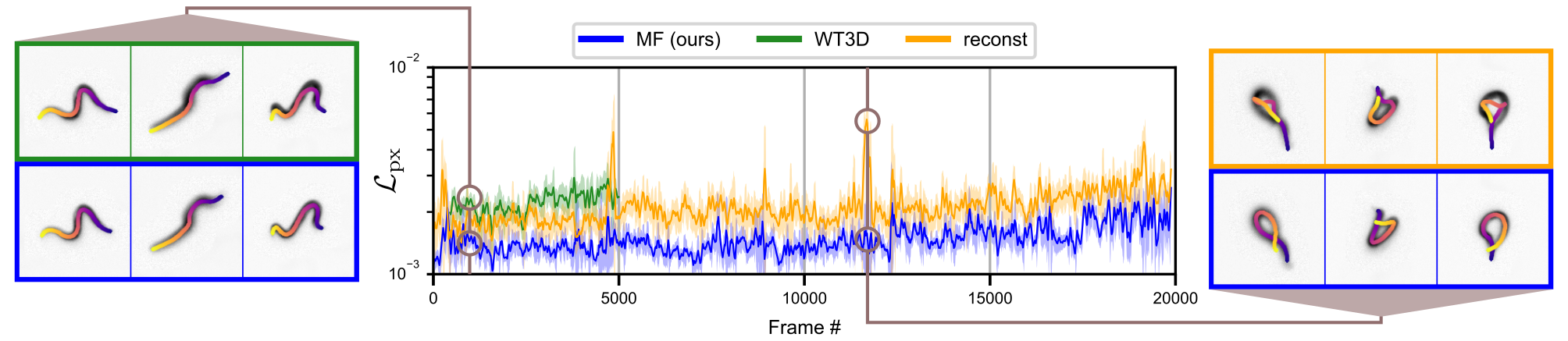}
	\caption{A comparison between our Midline Finder (MF), Yuval's Worm-Tracker 3D (WT3D) \cite{Omer2022} and Salfelder \etal's `reconst' \cite{Salfelder2021} methods across a single trial ($\simm \SI{13}{\minute}$). In the majority of cases our method generates midlines that better match the data (lower pixel losses, $\mathcal{L}_\text{px}$). We show moving averages over 25 frames ($\simm \SI{1}{\second}$) with shaded areas indicating $\pm 2 \text{std}$.}
	\label{fig:comparisons}
\end{figure*}
\begin{figure*}
	\centering
	\includegraphics[width=0.95\linewidth]{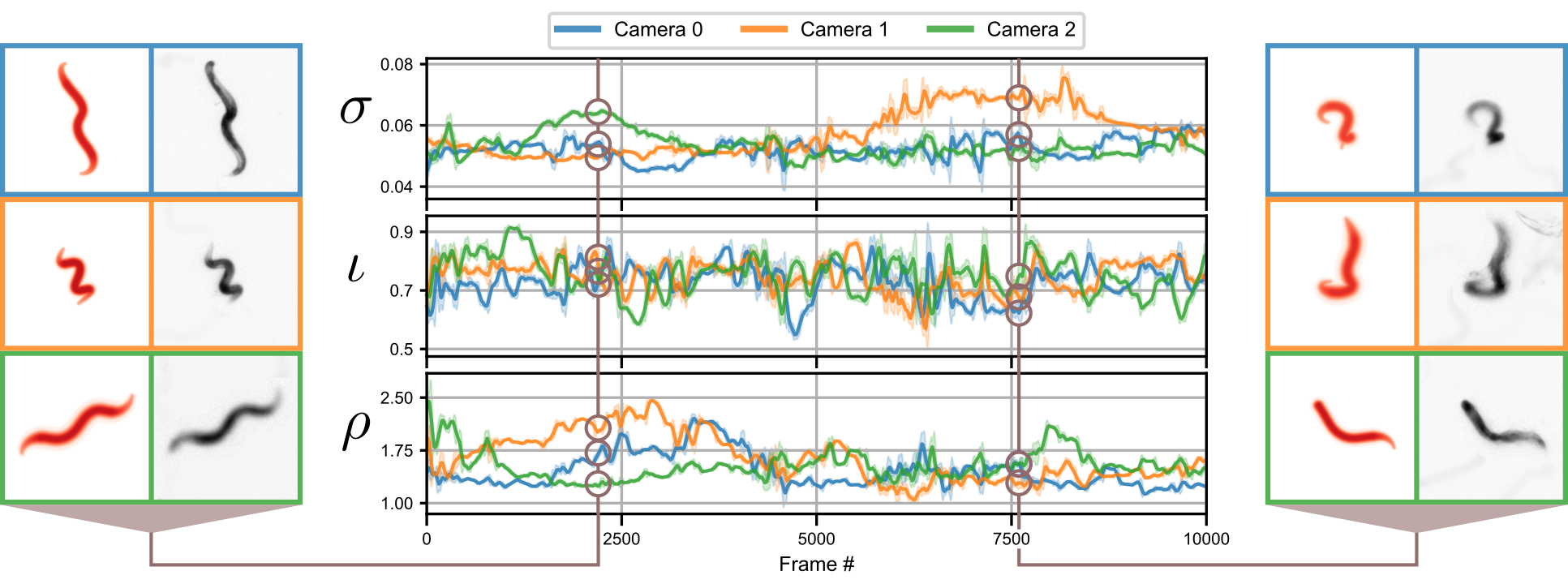}
	\caption{The rendering parameters change continually over the course of a recording to capture optical changes. Clear images (\eg early frames in cameras 0 and 1, switching to late frames in camera 2) are consistent with small values of $\sigma$ and large values of $\rho$. Blurry images (early camera 2, late camera 1) use high $\sigma$ and small $\rho$. We show moving averages over 25 frames ($\simm \SI{1}{\second}$) with shaded areas indicating $\pm 2 \text{std}$. Example comparisons between the renders (red) and raw images (grey) are shown on either side.}
	\label{fig:render_params}
\end{figure*}

Using our method we generate high quality 3D midline reconstructions for 43 of 44 recordings. One fails due to excessive coiling of the worm. Significant occlusions also occur during successful reconstructions and when combined with loss of focus can cause the shape to be lost. Video clips of good and poor reconstructions through challenging environmental conditions are described in \cref{app:results} along with ablation results to show benefits of each component.

We compare 2D reprojections of our midlines against 487 manual annotations that were produced from single images in isolation and contain a varying number of unordered points. We calculate the minimum distance from each annotated point to any reconstructed point and vice-versa and find that our midlines consistently come close ($\simm 2\text{px}$) to hand-annotated points (\cref{fig:validation}). Annotated points at the ends show an increased distance ($\simm 10\text{px}$) to our midline points. This shows that our curves generally fall short of reaching the very tips of the worm by $\sim \mathcal{O}(\text{worm radius})$.

Our method significantly outperforms previous methods developed using the same dataset \cite{Salfelder2021,Omer2022} when evaluated against the manual annotations (\cref{tab:validation_comparison}), but these only cover a selection of hand-picked examples. For a large-scale comparison we take 3D midlines and camera parameters found by each method and, using our pipeline, render them to generate comparable images (re-optimising the render parameters for their midlines, see \cref{sec:comparisons}). We skip the scoring and masking and calculate $\mathcal{L}_\text{px}$. The results (\cref{fig:comparisons}) show our method consistently produces shapes that more closely match the raw images. The biggest advantage over previous approaches is the improvement in robustness; we recover \SI{4}{\hour} \SI{37}{\minute} (ours) versus \SI{1}{\hour} \SI{32}{\minute} \cite{Salfelder2021} and \SI{45}{\minute} \cite{Omer2022}.

\cref{fig:render_params} shows the rendering parameters during a trial as the worm moves in and out of focus in the different cameras. Clearer images result in smaller values of $\sigma$ and larger values of $\rho$. The fluctuations in intensity $\iota$ are due in part to the posture of the worm in relation to the camera; when it is pointing directly towards the camera we see higher values of $\iota$ used to capture the darker image observed and when the shape is perpendicular to the camera we see lower values of $\iota$ to emulate the worm's transparency. All three parameters work in tandem to produce the final effect.

%------------------------------------------------------------------------

\section{Conclusion}

We present a robust and reliable framework for the 3D reconstruction of a microscopic, semi-transparent subject moving through a fluid and evaluate against two other algorithms and manually annotations. The key contribution of our approach -- constructing unique differentiable renderings for each view -- allows us to solve shape reconstruction and camera parameter optimisation by direct image comparison. This avoids feature extraction and correspondence matching, and hence offers a powerful alternative when those approaches are not well-suited, \eg due to the variation in appearance between views.

Multi-view microscopic camera calibration, imaging through fluids and parametric model fitting of semi-transparent subjects are challenges that have received little attention in the literature. While we have focused here on constructing a curve to fit a microscopic worm from three views, our method could be applied to the 3D reconstruction of arbitrary shape models at any scale using any number of viewpoints. Rendering points with adaptable super-Gaussian functions presents an effective solution to transparency and focal issues,  but more generally, our results indicate that our direct optimisation approach may offer an effective alternative to contemporary methods for 3D approximation of generic objects from a limited number of silhouette-like images.

%%%%%%%%% REFERENCES
{\small
\bibliographystyle{ieee_fullname}
\bibliography{Research}
}

\clearpage

\appendix

\renewcommand{\theequation}{S\arabic{equation}}
\renewcommand{\thetable}{S\arabic{table}}
\renewcommand{\thefigure}{S\arabic{figure}}

\section*{Supplementary material}

In this supplementary material we provide additional detail for some of the steps in our method and include the ranges and values for the various parameters and hyper-parameters. In \cref{sec:comparisons} we provide a more detailed comparison between our method and the two other methods available. We report the results of an ablation study in \cref{sec:ablations} and in \cref{app:results} we describe the three supporting videos (available here: \url{https://doi.org/10.6084/m9.figshare.22310650}) that demonstrate the method and showcase some of the results.

%------------------------------------------------------------------------

\section{Geometric model: The Bishop frame}
\label{app:bishop_frame}

3D curves are typically expressed in the Frenet frame $TNB$ where $T$ refers to the normalised tangent of the curve, $N$ is the `normal' vector defined as the normalised arc-length derivative of $T$ and $B$ is the `binormal' vector obtained through the cross product $B=T \times N$. This frame is defined along the curve using the Frenet-Serret formulas:
\begin{align}
\dot{T} &= \kappa N, \label{eq:frenet1} \\
\dot{N} &= -\kappa T + \tau B, \\
\dot{B} &= -\tau N,
\end{align}
where dot denotes the arc-length derivative $d/ds$ and $\kappa$ and $\tau$ are scalar fields generally called curvature and torsion respectively. For simplicity we leave the arc length parameter $s$ implicit in all equations.

A difficulty with the Frenet formulation is that the torsion, $\tau$, is strictly undefined for straight curves, or locally wherever $\kappa=0$. Zero (or near-zero) curvature is expected in an animal that propagates sinusoidal waves along its body and at these points we cannot guarantee a unique and consistent parametrisation. To overcome this ambiguity we use the Bishop frame \cite{Bishop1975}, given by $TM^1M^2$ where $T$ again refers to the normalised tangent of the curve and $M^1,M^2$ form an orthogonal basis. The Bishop equations define how the frame changes along the curve:
\begin{align}
\dot{T} &= m^1 M^1 + m^2 M^2, \label{eq:bishop1} \\ 
\dot{M}^1 &= -m^1 T, \label{eq:bishop2} \\
\dot{M}^2 &= -m^2 T, \label{eq:bishop3}
\end{align}
where $m^1,m^2$ are scalar fields analogous to $\kappa,\tau$ that express the curvature in the $M^1$ and $M^2$ directions respectively.

While the Bishop frame improves the zero-curvature problem, it does leave a degree of freedom in the choice of the initial value of $M^1$ ($M^1_{\mathrm{init}}$) that can point in any direction perpendicular to the initial tangent $T_{\mathrm{init}}$. Any rotation of $M^1_{\mathrm{init}}$ around $T_{\mathrm{init}}$ will result in a different $(m^1,m^2)$ representation of the curvature, but this rotation angle can easily be recovered and different representations subsequently aligned.

As Bishop describes in \cite{Bishop1975} (and expanded here for completeness) the two frames are related through their scalar field components. $\kappa$ can be recovered from $m^1,m^2$ using \cref{eq:frenet1,eq:bishop1} as:
\begin{align}
\kappa = \left\lvert \frac{dT}{ds} \right\rvert = \left\lvert m^1 M^1 + m^2 M^2 \right\rvert = \sqrt{(m^1)^2 + (m^2)^2}.
\end{align}
To recover the torsion $\tau$ that describes the rotation of the Frenet frame around $N$ let $\theta$ be the angle between $N$ and $M^1$, then
\begin{align}
N &= M^1 \cos{\theta} + M^2 \sin{\theta}, \label{eq:N2} \\
B &= - M^1\sin{\theta} + M^2 \cos{\theta}, \\
m^1 &= \kappa \cos{\theta}\text{ and} \\
m^2 &= \kappa \sin{\theta}.
\end{align}
Differentiating \cref{eq:N2} with respect to arc length and substituting from \cref{eq:bishop2,eq:bishop3} we have:
\begin{align}
\dot{N} &= 
        \dot{\theta} \left( - M^1 \sin{\theta} + M^2 \cos{\theta} \right) \\
        &\quad\quad + T \left(  m^1 \cos{\theta} - m^2 \sin{\theta} \right)
    \\ 
%\dot{\theta} \left( - M^1 \sin{\theta} + M^2 \cos{\theta} \right) + T \left(  m^1 \cos{\theta} - m^2 \sin{\theta} \right) \\
        &= \dot{\theta} B - \kappa T (\cos^2{\theta} + \sin^2{\theta}) \\
        &\implies \tau = \dot{\theta}.
\end{align}
Thus, in the words of Bishop, ``$\kappa$ and an indefinite integral $\int \tau ds$ are polar coordinates for the curve $(m^1,m^2)$''.

%------------------------------------------------------------------------

\section{Project: Pinhole camera model}
\label{app:pinhole_cams}

The imaging setup is modelled using a triplet of pinhole camera models with tangential and radial distortion \cite{Hartley2003}. A single pinhole camera model is used to project 3D points into an image plane using a perspective transformation. The 15 parameters required for each camera model are summarised in \Cref{tab:cam_params}. 
These are divided into intrinsic, extrinsic and distortion parameters. The intrinsic parameters are $(f_x,f_y,c_x,c_y)$, where $f_x$ and $f_y$ are the focal lengths and $(c_x, c_y)$ is a principal point usually set to the image centre. The extrinsic parameters -- angles $(\phi_0,\phi_1,\phi_2)$ and a translation vector $t$ -- define the extrinsic transformation $M=[R|t]$, where
\begin{align}
R &= R_z(\phi_0) R_y(\phi_1) R_x(\phi_2), \\
t &= \begin{pmatrix}
		t_0 \\ t_1 \\ t_2
	\end{pmatrix}
\end{align}
and $R$ is a rotation matrix composed of three axial rotations:
\begin{align}
R_z(\phi_0) &=  \begin{pmatrix}
            		\cos{\phi_0}  & -\sin{\phi_0}  & 0 \\
            		\sin{\phi_0}  & \cos{\phi_0}   & 0 \\
            		0             & 0              & 1
            	\end{pmatrix}, \\
R_y(\phi_1) &=  \begin{pmatrix}
            		\cos{\phi_1}  & 0              & \sin{\phi_1} \\
            		0             & 1              & 0 \\
            		-\sin{\phi_1} & 0              &\cos{\phi_0}
            	\end{pmatrix} \text{ and }\\
R_x(\phi_2) &=  \begin{pmatrix}
            		1             & 0              & 0 \\
            		0             & \cos{\phi_2}   & -\sin{\phi_2} \\
            		0             & \sin{\phi_2}   & \cos{\phi_2}
            	\end{pmatrix}.
\end{align}
The radial and tangential distortion coefficients, $(k_1, k_2, k_3)$ and $(p_1, p_2)$ respectively, complete the parametrisation.

Due to imperfections in the camera model (and possibly environmental vibrations of the setup), using fixed pinhole camera model parameters yields results with errors that vary over time and space (\ie the accuracy depends on the worm position). Allowing the camera parameters to change freely between frames resolves the reconstruction errors, but the problem becomes under-determined and introduces drift into both the camera and curve parameters resulting in incorrect tracking. To compensate for these errors without allowing full freedom of movement we use fixed pinhole camera parameters and introduce frame-dependent variables that emulate relative movement between the cameras, hence limiting drift and providing stable reconstructions.

To this end, the standard pinhole camera model is extended to include $\eta^s=(dx,dy,dz)$, relative shifts along the local coordinate axes (see \cref{sec:project}). These parameters approximate changes in the relative positions and rotations of the cameras by applying pixel translations after the perspective transformation. Without loss of generality $\eta^s$ can be limited to one direction per camera, thus capturing only relative shifts. The shifts used in camera index $c$ are given by $(s_x, s_y)_c$ where:
\begin{align}
    (s_x, s_y)_0 &= (dx, 0), \label{eq:shiftx} \\
    (s_x, s_y)_1 &= (0,  -dy), \text{ and } \label{eq:shifty} \\
    (s_x, s_y)_2 &= (0, dz). \label{eq:shiftz} 
\end{align}

For 3D object point $(X, Y, Z)$, the corresponding projected image point $(u, v)$ is generated using the following procedure (when $z \neq 0$):
\begin{align} \label{eq:pinhole_model}
	\begin{pmatrix} x \\ y \\ z \end{pmatrix} &= R \begin{pmatrix}X \\ Y \\ Z\end{pmatrix} + t, \\
	x' &= \frac{x}{z} + \frac{s_x}{f_x}, \label{eq:pinhole_shift_x} \\
	y' &= \frac{y}{z} + \frac{s_y}{f_y}, \label{eq:pinhole_shift_y} \\
	r^2 &= x'^2 + y'^2, \\
	k &= 1 + k_1 r^2 + k_2 r^4 + k_3 r^6, \\
	x'' &= k x' + 2 p_1 x' y' + p_2 (r^2 + 2 x'^2), \\
	y'' &= k y' + p_1 (r^2	+ 2 y'^2) + 2 p_2 x' y', \\
	\begin{pmatrix} u \\ v \end{pmatrix} &= \begin{pmatrix} f_x x'' + c_x \\ f_y y'' + c_y \end{pmatrix}.
\end{align}
Note the inclusion of the shift parameters in \cref{eq:pinhole_shift_x,eq:pinhole_shift_y}.

%------------------------------------------------------------------------

\section{Rendering parameters: Tapering}
\label{app:render_params}

The rendering stage generates super-Gaussian blobs at the projected image locations of each curve vertex ($n$). The shape of the blobs in camera $c$ depends on the optimisable parameters: the scale $\sigma_c$, the intensity $\iota_c$ and the exponent used in the Gaussian $\rho_c$. To capture the worm shape we taper the values of $\sigma_c$ and $\iota_c$ from their optimisable values along the middle $60\%$ down to fixed minimum values $\sigma_{\text{min}}$ and $\iota_{\text{min}}$ respectively at the ends. The tapered outputs $\bar{\sigma}_c, \bar{\iota}_c \in \mathbb{R}^N$ are calculated thus: 
\begin{align}
\bar{\sigma}_{c,n} &= 
\begin{cases}
	\sigma_{\text{min}} (1 - \frac{5n}{N}) + \sigma_c \frac{5n}{N} 	 					& 0 \leq n < N/5 \\
	\sigma_c								 											& N/5 \leq n < 4N/5 \\
	\sigma_c (1 - \frac{n-4N/5}{N-4N/5}) + \sigma_{\text{min}} \frac{n-4N/5}{N-4N/5} 	& 4N/5 \leq n < N,
\end{cases}
\end{align}
and
\begin{align}
\bar{\iota}_{c,n} &= 
\begin{cases}
	\iota_{\text{min}} (1 - \frac{5n}{N}) + \iota_c \frac{5n}{N} 						& 0 \leq n < N/5 \\
	\iota_c									 											& N/5 \leq n < 4N/5 \\
	\iota_c (1 - \frac{n-4N/5}{N-4N/5}) + \iota_{\text{min}} \frac{n-4N/5}{N-4N/5}		& 4N/5 \leq n < N.
\end{cases}
\end{align}
These values are used in \cref{eq:blobs}.

%------------------------------------------------------------------------

\section{Mask generation}
\label{app:masks}

The input images are masked to focus the pixel-errors to a single region, local to the predicted curve, that is consistent across all three views and excludes any interference that does not correspond to the same mass. The masks $M \in \mathbb{R}^{3 \times w \times w}$ are generated in a similar way to the renders $R$ (see \cref{section:render}), but with a few notable differences. First, the blobs $B$ are normalised and weighted by the relative scores,
\begin{equation}
B'_{c,n} = \frac{B_{c,n}}{\sum_{i,j} B_{c,n}} \cdot \hat{S}_n,
\end{equation}
then combined by taking the maximum values as before,
\begin{equation}
M'_{c,i,j} = \max \lbrace B'_{c,n,i,j} \rbrace_{n=0,\dots,N-1},
\end{equation}
and finally passed through a threshold:
\begin{equation}
M_{c,i,j} =
\begin{cases}
	1		& M'_{c,i,j} \geq \Theta \\
	0.2		& M'_{c,i,j} < \Theta.
\end{cases}
\end{equation}
For $\Theta=0$ we have $M=1$ everywhere and no masking occurs. For $\Theta \sim 1$ the mask shrinks around the blobs that correspond to the highest scoring vertices, making $M=0.2$ almost everywhere. In all our experiments we fix $\Theta=0.1$ as this appears to produce a good balance. Note that we do not completely exclude the remaining points, but just reduce their intensity. This allows some gradient to flow from outside the detection region which is especially important in the early stages when none of the curve may be intersecting the correct pixel mass. It is also important to detach the masks from the gradient computation at this stage otherwise the curve will simply shrink and fade away from the high-intensity pixels thus minimising pixel errors simply by detecting fewer pixels.

%------------------------------------------------------------------------

\section{Centre-shifting}
\label{app:centre_shift}

The curve is periodically shifted along its length to centre it over the pixel mass in all three views. An unbalanced alignment can be seen from the score profile when the centre-of-mass index of $\hat{S}$ ($\bar{n}$) is not in the middle of the curve (\ie $\bar{n} \neq N/2$). We can then shift the curve along its length using $\bar{n}$ as the new midpoint, removing vertices from the low-scoring end and adding new vertices to the high-scoring end. The low-scoring end will consequently improve, and since there is no expectation that the new vertices will match the images this typically means the high-scoring end worsens; rectifying the imbalance.

To perform a centre-shift we calculate the centre of mass of the score profile and the degree of imbalance as:
\begin{align}
\bar{n} &= \frac{\sum_n n \hat{S}_n }{\sum_n \hat{S}_n}, \text{ and} \\
n_s &= \bar{n}-N/2.
\end{align}
Then we update the curvature by shifting the values and decreasing linearly to zero at the ends. I.e. for $n_s > 0$,
\begin{equation}
K_n \leftarrow
\begin{cases}
	K_{n+n_s}		& 0 \leq n < N-n_s, \\
	K_{N - n_s - 1} (1 - \frac{N - n_s - n + 1}{n_s})		& N-n_s \leq n < N,
\end{cases}
\end{equation}
and similarly for $n_s < 0$. Finally, new position and orientation parameters are calculated from the adjusted midpoint and updated curvatures using \cref{eq:curve}:
\begin{equation}
(P,T,M^1) \leftarrow F\left(P_{\bar{n}},T_{\bar{n}},M^1_{\bar{n}},K,l,\bar{n}\right).
\end{equation}
This process is illustrated in \cref{fig:centre_shift}. 

Centre-shifting the curve occurs between gradient descent optimisation steps. In practise, shifting after every step quickly leads to instabilities as the new points are not afforded the time required to align them with the images. Furthermore, it is unrealistic to expect a perfect balance can be sustained and an unconstrained $n_s$ means large shifts may be applied, possibly due to a change in the camera parameters or some transient interference, that would destroy extensive sections of the curve. To mitigate these problems we only apply centre-shifting every $\alpha$ steps when $|n_s| > \beta N$ and then restrict the shift size to $\gamma$ (\ie $n_s \leftarrow \min \lbrace n_s, \gamma \rbrace$). In our experiments we find values of $\alpha \in [3,6]$, $\beta \in (0.05,0.1)$ and $\gamma \in [1,2]$ provide the necessary stabilisations (\Cref{tab:parameters}).

%------------------------------------------------------------------------

\section{Optimisation}

The non-optimisable parameter values and ranges that are used in our experiments are outlined in \Cref{tab:parameters}. The biggest factors affecting the choice of parameters are the magnification and individual worm size -- both of which vary between experiments. These determine the required image size, $w$, and inform the estimates for the length bounds, $l_\text{min}$ and $l_\text{max}$. The super-Gaussian blobs are generated in corresponding $w\times w$ images, so the minimum scales and intensities at the tips, $\sigma_\text{min}$ and $\iota_\text{min}$, must also change accordingly with the image and worm size. 

\Cref{tab:omegas} lists the weighting coefficients used in the combined loss calculation (\cref{eq:loss}). Values of $\omega_\text{sm}$ and $\omega_\text{t}$ may vary between experiments to capture the different dynamics observed in the different environmental conditions (specifically, concentration of the gelatin). For example, when the worm is deforming quickly (in low-viscosity experiments) there are large postural changes between frames and therefore the temporal loss $\mathcal{L}_\text{t}$ is relatively big. In this case a smaller value of $\omega_\text{t}$ is used to prevent the reconstruction lagging behind the worm. Similarly, when the worm is deforming slowly (in high-viscosity experiments) it frequently forms tightly coiled postures in which case the smoothness loss $\mathcal{L}_\text{t}$ is large and a smaller value for $\omega_\text{sm}$ is more suitable. As discussed in the main text, when reconstructing full sequences the initial curve and parameters are used for the initial guesses to the subsequent frame. This preserves head-tail orientation and consistency through complex manoeuvres. 

The learning rates are shown in \Cref{tab:lrs} -- these are fixed for all experiments. Optimisable camera, curve and render parameters are summarised in \Cref{tab:cam_params,tab:curve_params,tab:render_params} respectively.

%------------------------------------------------------------------------

\section{Comparisons with previous methods} 
\label{sec:comparisons}

\begin{figure}
	\centering
	\includegraphics[width=0.95\linewidth]{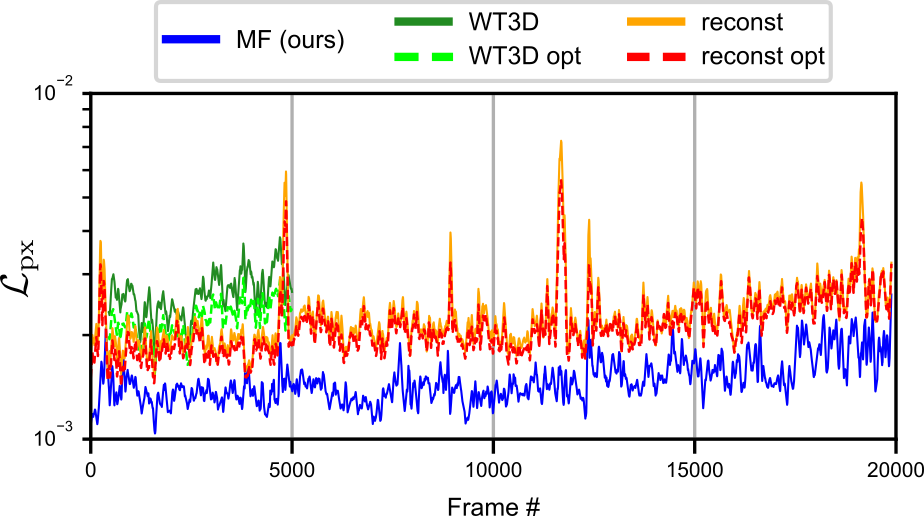}
	\caption{A comparison between the losses obtained when using our renderer parameters for Yuval's Worm-Tracker 3D (WT3D) \cite{Omer2022} and Salfelder \etal's `reconst' \cite{Salfelder2021} midlines \vs the losses obtained after re-optimising the renderer parameters to better suit their midlines. Our method (MF) is shown for reference. We show moving averages over 25 frames (\SI{1}{\second}).}
	\label{fig:comparisons_opt}
\end{figure}

In \cref{fig:comparisons} we compare results against two existing methods \cite{Salfelder2021,Omer2022}. These methods only provide projected midline coordinates, so in order to use our pipeline to generate renders and calculate $\mathcal{L}_\text{px}$ we need to provide renderer (blob) parameters. We can use values found for our midlines, but this introduces a bias towards our method as these parameters are only optimal for our midlines. To mitigate against this we initialise with our values but re-optimise for each frame until convergence (keeping the curve and camera parameters fixed) to ensure optimal rendering parameters are found for each midline. In \cref{fig:comparisons_opt} we show the effect of the re-optimisation across the same clip. As expected, re-optimisation reduces the loss for both methods, but the improvement is fairly marginal. The improved losses are used for comparison in the main text.

In \Cref{tab:validation_comparison} we evaluate midline quality of all three methods against ground-truth manual annotations. This unbiased evaluation (as only projected midline points are used) shows our method to be more accurate. However, the sample size is limited, more so as the other methods only provide results for roughly half of the available annotations.

\begin{table}[ht]
    \centering
    \begin{tabularx}{\linewidth}{LLL|L}
        \textbf{Method} & \textbf{\# poses} & \textbf{Score} & \textbf{Total} \\
        \midrule 
        MF (ours) & 487 & \textbf{1.53} (0.51) & \textbf{\SI{4}{\hour} \SI{37}{\minute}} \\
        \midrule 
        reconst \cite{Salfelder2021} & 226 & 1.54 (0.69) & \SI{1}{\hour} \SI{32}{\minute} \\
        MF (ours) & 226 & \textbf{1.34} (0.53) & \\
        \midrule 
        WT3D \cite{Omer2022} & 237 & 2.64 (0.92) & \SI{45}{\minute} \\
        MF (ours) & 237 & \textbf{1.46} (0.59) & \\
        \bottomrule
    \end{tabularx}
    \caption{
    Mean (and standard deviation) pixel distances between predicted points and hand-annotated points (see \cref{fig:validation}). Total refers to the overall reconstructed duration using each method. 
    }
    \label{tab:validation_comparison}
\end{table}

%------------------------------------------------------------------------

\section{Ablation study}
\label{sec:ablations}

We demonstrate some of the effects of masking (\cref{fig:masks}), centre-shifting (\cref{fig:centre_shift}) and varying parameters (Movie 1) throughout the main text. In \Cref{tab:ablation} we present the results from a more thorough ablation study conducted over a typical $\sim$\,\SI{5}{\minute} clip to clarify the importance of the different components of our method. 

Optimisable camera (a), rendering parameters (b) and centre-shifting (c) yield considerable benefit; setting \mbox{$\omega_\text{sc}>0$} (d) and regularisation (f) incur a marginal cost ($\tilde{\mathcal{L}}> 0.99$) but recover poses with high transparency (d) and ensure realistic (smooth) poses (f); finally, masking (e) incurs a small cost but provides robustness to sequences with interference (\eg Movie 1).

\begin{table}[ht]
    \centering
    \begin{tabularx}{\linewidth}{lLL}
        \textbf{Variant} & $<\mathcal{L}_\text{px}>$ & $\tilde{\mathcal{L}} = <\mathcal{L}_\text{px}>/\text{ Ref}$ \\
        \midrule 
		Ref & \num{3.99e-03} (\num{1.13e-03}) & 1.000 \\
        \midrule        
		a & \num{7.21e-03} (\num{1.97e-03}) & 1.807 \\
		b & \num{4.64e-03} (\num{1.42e-03}) & 1.163 \\
		c & \num{4.29e-03} (\num{1.71e-03}) & 1.075 \\
		d & \num{3.99e-03} (\num{1.12e-03}) & 1.000 \\
		e & \num{3.69e-03} (\num{9.00e-04}) & 0.925 \\
		f & \num{3.96e-03} (\num{1.08e-03}) & 0.992 \\
        \bottomrule
    \end{tabularx}
    \caption{
    Ablation results (mean (and standard deviation) pixel errors and normalised mean) across a typical 7200-frame clip. Variants: 
    a) no camera parameter optimisation, b) rendering parameters ($\sigma,\iota,\rho$) fixed to averages from the reference results, c) no centre-shifting, d) no scores-loss ($\omega_\text{sc}=0$), e) no input masking, f) no regularisation losses ($\omega_\text{sm}=\omega_\text{t}=0$).
    }
    \label{tab:ablation}
\end{table}

%------------------------------------------------------------------------

\section{Supplementary results} \label{app:results}

Three supporting videos are available here: \url{https://doi.org/10.6084/m9.figshare.22310650}.

In the accompanying Movie 1, the effects of adjusting some of the parameters listed in \Cref{tab:parameters,tab:omegas} on the solution are demonstrated. In this video we show the optimisation process using paired examples. The same frame and randomised initial guess are used for each pair and the optimisation is run for a fixed 2000 steps. The frames are selected to demonstrate a range of challenging conditions -- especially for achieving convergence from a random guess. The first of each pair shows successful optimisation using parameter values in the ranges specified in \Cref{tab:parameters,tab:omegas}. The second of each pair shows the effect that changing one of the parameters has on the converged solution. 

We include two further videos demonstrating examples of both successful and less-successful sequence reconstructions. In Movie 2 we showcase successful examples. First, when the worm is well resolved in all three views. Next, when there is interference from dirt or bubbles and/or poor focus in one or more views. Lastly, through complex coiling manoeuvres that include significant self-occlusion.

The limitations of our method are illustrated in Movie 3. These examples, taken from otherwise successful reconstructions, demonstrate that when significant loss of focus is combined with coiling, heavily occluded, postures the reconstruction can fail. The exact degree of failure is difficult to ascertain for the exact same reasons and only by watching the full sequences can we be convinced that the reconstruction is incorrect. This suggests that incorporating more temporal information may help to resolve these fail-cases, but we leave this for future investigation.

\begin{table*}
    \centering
    \begin{tabular}{m{0.1\textwidth} m{0.6\textwidth} m{0.2\textwidth}}
        \textbf{Parameter} & \textbf{Purpose} & \textbf{Domain} \\
        \midrule 
        $f_x$,$f_y$ & Focal lengths & $\mathbb{R}^+$ \\
        $(c_x, c_y)$ & Principal image point & $\mathbb{R}^+$ \\
        $\phi_0,\phi_1,\phi_2$ & Rotation angles & $[0, 2\pi)$ \\
        $t$ & Translation (position) vector & $\mathbb{R}^3$ \\
        $(k_1, k_2, k_3)$ & Radial distortion coefficients & $\mathbb{R}^+$ \\
        $(p_1, p_2)$ & Tangential distortion coefficients & $\mathbb{R}^+$ \\
        $\eta^s$ & Relative shifts & $\mathbb{R}^3$ \\
        \bottomrule
    \end{tabular}
    \caption{Camera model parameters. With the exception of $\eta^s$ these are defined for each camera. $\eta^s$ is shared between the models as per \cref{eq:shiftx,eq:shifty,eq:shiftz}.}
    \label{tab:cam_params}

    \vspace{2em}
    \begin{tabular}{m{0.1\textwidth} m{0.6\textwidth} m{0.2\textwidth}}
        \textbf{Parameter} & \textbf{Purpose} & \textbf{Domain} \\
        \midrule 
        $P$ & 3D curve vertex coordinates & $\mathbb{R}^{N \times 3}$ \\
        $T$ & Normalised curve tangent vectors at each vertex location & $\mathbb{R}^{N \times 3}$ \\
        $M^1$ & Normalised curvature orientation vectors at each vertex location & $\mathbb{R}^{N \times 3}$ \\
        $K$ & Vector curvature & $\mathbb{R}^{N \times 2}$ \\
        $l$ & Curve length & $(l_\text{min}, l_\text{max})$ \\
        \bottomrule
    \end{tabular}
    \caption{Curve and Bishop frame parameters.}
    \label{tab:curve_params}

    \vspace{2em}
    \begin{tabular}{m{0.1\textwidth} m{0.6\textwidth} m{0.2\textwidth}}
        \textbf{Parameter} & \textbf{Purpose} & \textbf{Domain} \\
        \midrule 
        $\sigma_c$ & Standard deviation of the super-Gaussian blobs along the untapered middle 60\% of the worm in camera $c$ & $[\sigma_\text{min}, \infty)$ \\
        $\iota_c$ & Intensity scaling factor for the super-Gaussian blobs along the untapered middle 60\% of the worm in camera $c$ & $[\iota_\text{min}, \infty)$ \\
        $\rho_c$ & Exponent used in the super-Gaussian blobs in camera $c$ & $(0, \infty)$ \\
        \bottomrule
    \end{tabular}
    \caption{Rendering parameters.}
    \label{tab:render_params}
\end{table*}

\begin{table*}[t]
    \centering
    \begin{tabular}{m{0.1\textwidth} m{0.6\textwidth} m{0.2\textwidth}}
        \textbf{Parameter} & \textbf{Purpose} & \textbf{Value/Range} \\
        \midrule 
        $w$ & (Square) image size & $200$--$350~\si{px}$ \\
        $N$ & Number of discrete curve vertices & $128$ \\
        $l_\text{min}$ & Minimum curve length & $0.5$--$1~\si{mm}$ \\
        $l_\text{max}$ & Maximum curve length & $1$--$2~\si{mm}$ \\
        $k_\text{max}$ & Maximum curvature constraint & $3$ osculating circles \\
        $\sigma_\text{min}$ & Standard deviation of the super-Gaussian functions at the tips & $2$--$4~\si{px}$ \\
        $\iota_\text{min}$ & Intensity scaling factor of the super-Gaussian functions at the tips & $0.15$--$0.3$ \\
        $\Theta$ & Mask threshold & $0.1$ \\
        $\alpha$ & Frequency of centre-shift adjustments (number of gradient descent steps) & $3$--$6$ steps \\
        $\beta$ & Centre-shift adjustment sensitivity & $0.05$--$0.1$ \\
        $\gamma$ & Maximum centre-shift adjustment & $1$--$2$ vertices \\
        \bottomrule
    \end{tabular}
    \caption{Non-optimisable parameter values and ranges used in our experiments. Listed in the order they appear in the text.}
    \label{tab:parameters}
    
    \vspace{2em}
    
    \centering
    \begin{tabular}{m{0.1\textwidth} m{0.6\textwidth} m{0.2\textwidth}}
        \textbf{Parameter} & \textbf{Purpose} & \textbf{Value/Range} \\
        \midrule 
        $\omega_\text{px}$ & Weighting of the pixel loss $\mathcal{L}_\text{px}$ & $0.1$ \\
        $\omega_\text{sc}$ & Weighting of the scores loss $\mathcal{L}_\text{sc}$ & $0.01$ \\
        $\omega_\text{sm}$ & Weighting of the smoothness loss $\mathcal{L}_\text{sm}$ & $10$--$100$ \\
        $\omega_\text{t}$ & Weighting of the temporal loss $\mathcal{L}_\text{t}$ & $10$--$100$ \\
        $\omega_\text{i}$ & Weighting of the intersections loss $\mathcal{L}_\text{i}$ & $0.1$--$1$ \\
        \bottomrule
    \end{tabular}
    \caption{Weighting coefficients for the different loss terms.}
    \label{tab:omegas}
    
    \vspace{2em}
   
    \centering
    \begin{tabular}{m{0.1\textwidth} m{0.6\textwidth} m{0.2\textwidth}}
        \textbf{Parameter} & \textbf{Purpose} & \textbf{Value} \\
        \midrule 
        $\lambda_p$ & Learning rate for the curve parameters $\lbrace P, T, M^1, K, l \rbrace$ & $\num{1e-3}$ \\
        $\lambda_r$ & Learning rate for the rendering parameters $\lbrace \sigma, \iota, \rho \rbrace$ & $\num{1e-4}$ \\
        $\lambda_\eta$ & Learning rate for the camera parameters $\eta$ & $\num{1e-5}$ \\
        $\lambda_\text{min}$ & Minimum learning rate for all parameters & $\num{1e-6}$ \\
        \bottomrule
    \end{tabular}
    \caption{Learning rates for the different parameter groups.}
    \label{tab:lrs}
\end{table*}

\end{document}